%% file: acl2020.tex
\documentclass[11pt,a4paper]{article}
\usepackage[hyperref]{acl2020}
\usepackage{times}
\usepackage{latexsym}

\usepackage{microtype}
\usepackage{comment}
\usepackage{amsmath,amssymb,amsthm}
\usepackage{url}
\usepackage{graphicx}
\usepackage{subcaption,siunitx,booktabs}
\usepackage{tabularx}
\usepackage{makecell}
\newcommand{\ours}{ESPRIT}

\DeclareMathOperator*{\softmax}{softmax}
\aclfinalcopy % Uncomment this line for the final submission
\def\aclpaperid{150} %  Enter the acl Paper ID here

\usepackage[flushmargin,splitrule]{footmisc}
\title{\ours: Explaining Solutions to Physical Reasoning Tasks}
\author{
Nazneen Fatema Rajani$^{1*}$
\quad Rui Zhang$^{2*}$
\quad Yi Chern Tan$^2$
\\{\bf
\quad Stephan Zheng$^1$
\quad Jeremy Weiss$^2$
\quad Aadit Vyas$^2$
\quad Abhijit Gupta$^2$}
\\{\bf
\quad Caiming Xiong$^1$
\quad Richard Socher$^1$
\quad Dragomir Radev$^{1,2}$}\\
\\
$^*$Equal contribution. \\
$^1$ Salesforce Research
\quad  $^2$ Yale University\\
\tt{\{nazneen.rajani, stephan.zheng, cxiong, rsocher, dradev\}@salesforce.com}\\
\tt{\{r.zhang, yichern.tan, dragomir.radev\}@yale.edu}\\
}

\date{}

\begin{document}
\maketitle

\begin{abstract}
Neural networks lack the ability to reason about qualitative physics and so cannot generalize to scenarios and tasks unseen during training. We propose \ours, a framework for commonsense reasoning about qualitative physics in natural language that generates interpretable descriptions of physical events. 
We use a two-step approach of first identifying the pivotal physical events in an environment and then generating natural language descriptions of those events using a data-to-text approach.
Our framework learns to generate explanations of how the physical simulation will causally evolve so that an agent or a human can easily reason about a solution using those interpretable descriptions. Human evaluations indicate that \ours~produces crucial fine-grained details and has high coverage of physical concepts compared to even human annotations. Dataset, code and documentation are available at \url{https://github.com/salesforce/esprit}.
\end{abstract}

\input{01introduction.tex}
\input{02dataset.tex}
\input{03task_methods.tex}
\input{04evaluation_metrics.tex}
\input{05results_discussion.tex}
\input{06related_work.tex}
\input{07conclusion.tex}
\input{08acknowledgment.tex}

\bibliography{acl2020}
\bibliographystyle{acl_natbib}
\end{document}

%% file: 01introduction.tex
\section{Introduction}
Humans learn to understand and reason about physical laws just by living in this world and doing everyday things. AI models, on the other hand, lack this ability and so are unable to generalize to new scenarios that require reasoning about abstract physical concepts like gravity, mass, inertia, friction, and collisions \citep{bakhtin2019phyre}. We propose Explaining Solutions to Physical ReasonIng Tasks (\ours), a framework for explaining qualitative physics reasoning using natural language. Neural networks with knowledge of qualitative physics would have commonsense reasoning abilities about the way the world works ~\citep{forbus1988qualitative}. In turn, this could, for example, improve performance on tasks that involve interacting with humans and make human-robot interactions more efficient and trustworthy.

\begin{figure}[t!]
  \centering
  \includegraphics[width=0.45\textwidth]{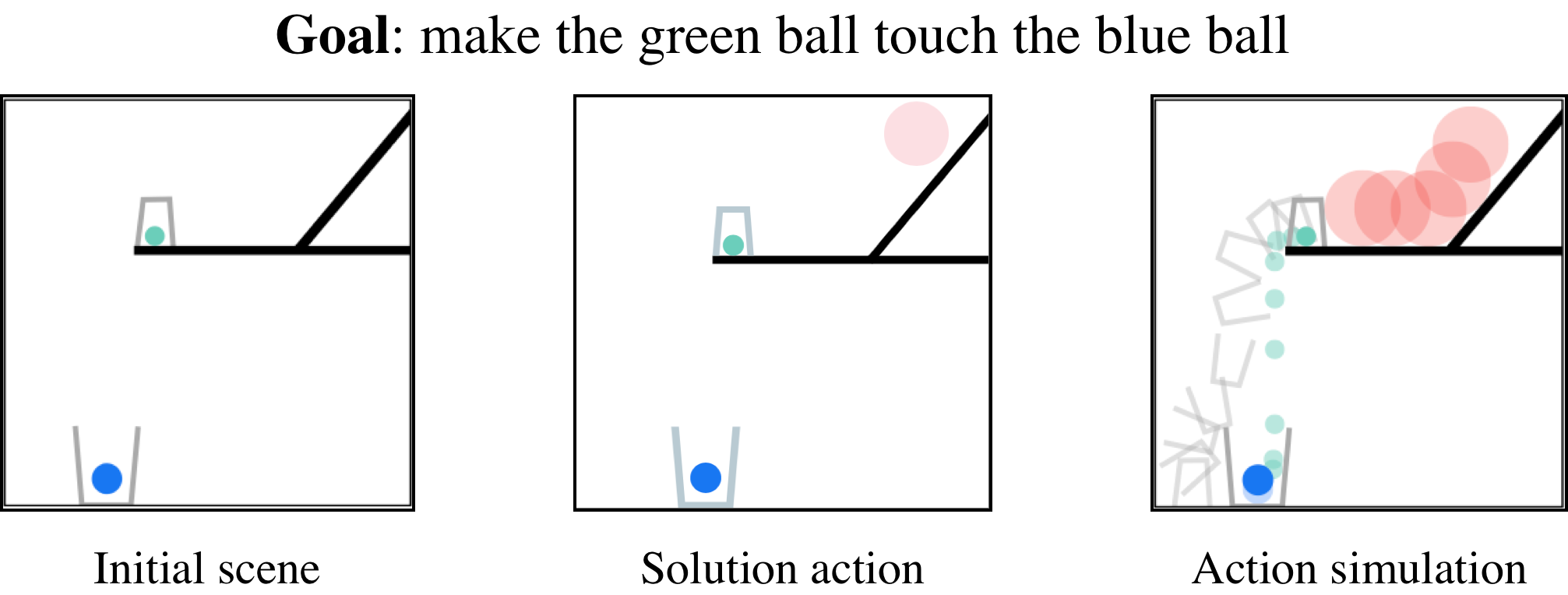}
  \caption{An example from the PHYRE dataset \citep{bakhtin2019phyre} consisting of a goal, an initial scene, a solution -- the action of adding a red ball, and the resulting simulation rollout.
  Each object color corresponds to an object type.
  Red: user-added dynamic object; Green and Blue: dynamic goal object; 
%   Purple (not pictured): static goal object; 
Gray: dynamic scene object; Black: static scene object.}
  \label{fig:phyre}
\end{figure}

Ideally, AI systems would reason about and generate natural language commonsense explanations of physical concepts that are relevant to their behavior and prediction. A key intuition is that natural language can provide an efficient low-dimensional representation of complicated physical concepts. To equip AI systems with this ability, we collected a set of open-ended natural language human explanations of qualitative physics simulations. The explanations include descriptions of the initial scene, i.e., before any physics is at play, and a sequence of identified pivotal events in a physics simulation. Three physical concepts are crucial for our simulation to reach a specified goal state: gravity, collision, and friction.

\begin{figure*}[!ht]
  \centering
  \includegraphics[width=\textwidth]{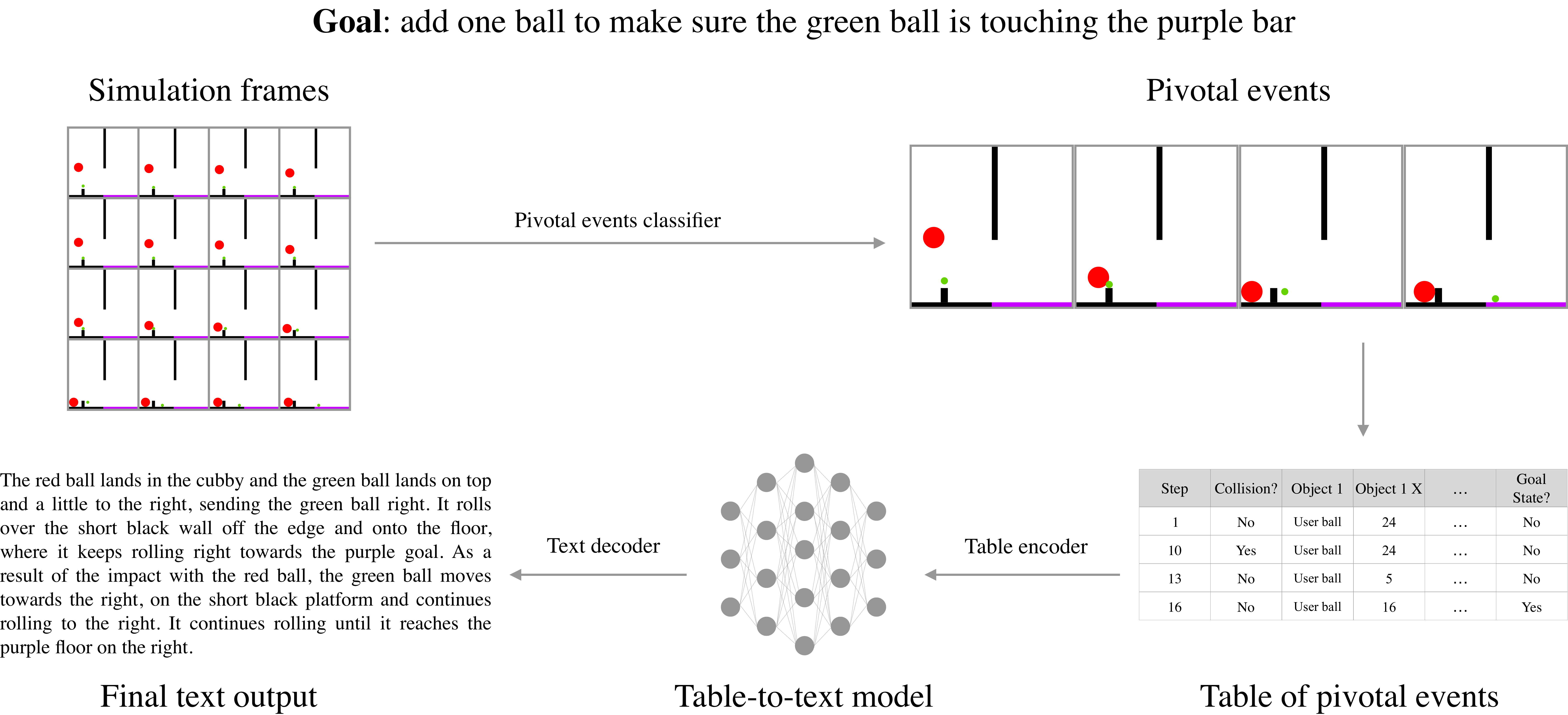}
  \caption{The end-to-end \ours~framework for identifying pivotal physical events, extracting the features from pivotal events in a table, and explaining solutions using a table-to-text model for natural language generation. The purple bar is a static goal object.}
  \label{fig:esprit}
\end{figure*}

Our work attempts to build an interpretable framework for qualitative physics reasoning with strong generalization abilities mirroring those of humans. \ours~is the first-ever framework that unifies commonsense physical reasoning and interpretability using natural language explanations. Our framework consists of two phases: (1) identifying the pivotal physical events in tasks, and (2) generating natural language descriptions for the initial scene and the pivotal events. In the first phase, our model learns to classify key physical events that are crucial to achieving a specified goal whereas in the second phase, our model generates natural language descriptions of physical laws for the events selected in the first phase. We demonstrate \ours~on the PHYsical REasoning (PHYRE) benchmark \citep{bakhtin2019phyre}.
PHYRE provides a set of physics simulation puzzles where each puzzle has an initial state and a goal state.
The task is to predict the action of placing one or two bodies (specifically, red balls of variable diameters) in the simulator to achieve a given goal. Figure~\ref{fig:phyre} shows an example of a task with a specified goal.

The input to \ours~is a sequence of frames from a physics simulation and the output is a natural language narrative that reflects the locations of the objects in the initial scene and a description of the sequence of physical events that would lead to the desired goal state, as shown in Figure~\ref{fig:esprit}. The first phase of the framework uses a neural network classifier to identify salient frames from the simulation. 
For the second phase we experimented with table-to-text models \citep{puduppully2019data, puduppully2019entity} as well as pre-trained language models \citep{radford2018improving}. We evaluated our framework for natural language generated reasoning using several automated and human evaluations with a focus on the understanding of qualitative physics and the ordering of a natural sequence of physical events. We found that our model achieves very high performance for phase one (identifying frames with salient physical events) and that, for phase two, the table-to-text models outperform pre-trained language models on qualitative physics reasoning. 
%Note that we do not use the actual image representation of the simulation for any of our models.

%% file: 02dataset.tex
\section{Dataset}
\subsection{PHYRE Benchmark}
We build our dataset by extending PHYRE \citep{bakhtin2019phyre}, a recent benchmark dataset for PHYsical REasoning.\footnote{\url{https://phyre.ai/}}
PHYRE consists of a set of physics puzzles in a simulated 2D environment.
This environment follows simple deterministic Newtonian physics with a constant downward gravitational force and a small amount of friction.

All objects (balls, bars, standing sticks, and jars) are non-deformable, and each object color corresponds to an object type: red is the user-added dynamic object; green and blue are used for dynamic objects that are part of the goal state; purple is for static goal objects; gray is for dynamic scene objects; black is for static scene objects.
Each task starts with an initial scene and has a goal state, described in natural language.
The task can be solved by placing one or two red balls in the simulation environment and choosing their sizes in a way that when the simulation runs according to the laws of physics the goal state is achieved.
No further action can be taken after the simulation starts.

In this paper, we focus on the $25$ task templates in the PHYRE dataset that involve the placement of a single ball to reach the goal state.
Each template defines a set of $100$ similar tasks generated by using different parameters for a template such as positions and sizes of objects.
All tasks within the same template have the same goal (e.g., ``make the blue ball touch the green ball") but somewhat different initial configurations.

\begin{table}[t!]
\centering
\resizebox{\columnwidth}{!}{
\begin{tabular}{lr}
\toprule
Templates                          & 25     \\
Tasks                              & 2441   \\
Train/Val/Test                     & 1950/245/246   \\ 
Objects / Task                     & 14   \\ 
Frames / Task                      & 658  \\ 
Events / Task                      & 54   \\ 
Salient Events / Task              & 7    \\ 
Tokens / Initial State Description & 36     \\
Tokens / Simulation Description    & 45     \\
Vocabulary Size                    & 2172   \\
\bottomrule
\end{tabular}
}
\caption{Statistics for the \ours~Dataset.}
\label{tab:data}
\end{table}

\subsection{Representing Frames as Structured Tables}
\label{sec:representation}
We represent the simulation frames as structured tables by extracting information using the simulator module in the PHYRE API.\footnote{\url{https://phyre.ai/docs/simulator.html}} The simulations consist of $60$ frames per second.
For each object, we collect its id, type (boundary, bar, jar, circle), color (red, green, blue, purple, gray, black), state (dynamic, static), and $(x,y)$ coordinates. Jars also have an angle of rotation,  width,  base length, and side length (referred to as just “length”). Bars have length, width, and angle of rotation while circles have a radius.
For each collision between two objects, we collect the $(x,y)$ coordinates, velocity as a $( v_x, v_y )$ vector, and the angle of rotation in radians for each object involved in the collision.

\paragraph{Extracting data from the PHYRE simulator.}
To track the motion of objects through a simulation, we intercepted PHYRE's built-in simulator.
First, we created a dictionary of objects and their attributes in the simulation's initial scene (including the predicted action that was performed). It is important to note that the dictionary contains properties of both static and dynamic objects. But because static objects such as the simulation boundary are not affected by the physics in the simulation and their properties never change. So, unless a static object is involved in a collision, we did not collect any other data about that object during the simulation.

Once this initial pass was made, we extracted the images of frames generated for the $2500$ single-ball simulations. Each simulation was run for a maximum of $1000$ time steps or approximately $16$ seconds. After the initial action is taken, a simulation is considered successful if it reaches the goal state and remains in that state for at least 180 consecutive time steps, the equivalent of three seconds. If a simulation does not satisfy this goal condition, it is considered unsuccessful.
In this way, we found solution simulations for $2441$ out of $2500$ tasks.
The remaining $59$ task simulations seem more complex and would possibly require a prohibitive number of trials ($>10000$) to reach the goal successfully and so we excluded those from our dataset.

Finally, we mapped the dictionary of objects and attributes in the initial state to the frames derived from the simulator so that we could track how the object's properties change from one frame to another.

\paragraph{Generating tables.} The three physical concepts at play in the simulations -- friction, collision, and gravity are either a cause or an effect of some collision. Therefore, collisions were the most common physical event in the simulations (average = $54$ per task) and so we decided to only record collisions. 
For every collision extracted, we applied a window of size $3$ to fetch frames before and after the collisions to remove any noise and get the more precise timestamp of the collision. Because pivotal events in a solution simulation only occur when two objects collide or separate, like a ball falling onto another or a ball rolling off of an elevated bar, we treat both cases identically.

\subsection{Two-stage Annotation Procedure}
Based on the simulation screenshots of the initial state and the collision, we employed a two-stage annotation procedure using Amazon MTurk.
In the first stage, we showed the goal, the initial state, and all collisions during the simulation.
We asked annotators to pick pivotal or salient events by selecting all and only the collisions that are {\it causally} related to the placement of the red ball and are necessary for the completion of the goal.
In the second stage, we collected human annotations of natural language descriptions for the initial scene and explanations for the sequence of salient collisions annotated during the first stage. We showed the annotators the goal, the initial state with the red ball added, an animated GIF of the simulation, and the frames of salient collisions. We asked them to include descriptions of the shape, color, and position of the objects involved. The annotations for the initial scene and salient collisions are collected in separate text boxes.

\subsection{Data Statistics}
Our data statistics are summarized in Table \ref{tab:data}.
We generated solutions for $2441$ tasks, covering $25$ different templates.
These tasks have an average of $14$ objects, $658$ total frames, and $54$ collision events.
We split the tasks randomly into $1950$ train, $245$ validation, and $246$ test.
On average, each task has $7$ events marked as salient by the annotators.
Also, on average the description of the initial state and simulation each have about $40$ tokens, with a vocabulary size of $2172$.

%% file: 03task_methods.tex
\section{Tasks and Methods}
\label{sec:methods}
\ours{} includes the following components:
\paragraph{Pivotal event detection.}
Given all the collision events in the simulation, select collisions that are crucial to achieving the goal state.
Pivotal or salient collisions are collisions that fulfill the following two criteria: (i) causally related to the placement of the red ball, and (ii) necessary for the completion of the given goal.

To train a classifier to detect salient events, we use the following features from the table representation: collision time step, each object's shape, position $(x,y)$, velocity $(v_x,v_y)$, and angle of rotation. This totals $13$ input features. The first object is often static, such as the boundary, while the second is often dynamic, such as the user-placed red circle.
We experimented with a decision tree and a neural network MLP classifier to compare with a baseline that classifies every frame as salient.
The MLP has three layers with $128$, $128$, and $32$ nodes. There is a $15\%$ dropout to avoid overfitting and batch normalization between each layer. Finally, a sigmoid node converts the output into a probability from $0$ to $1$ (anything above $50\%$ is classified as salient).
The models are trained on $59179$ collisions ($52066$ negative, $7113$ positive) and tested on $6893$ collisions ($6000$ negative, $893$ positive).
\paragraph{Natural language description of initial states.}
Given a list of objects and their attributes (color, position, type) in the initial frames, generate a corresponding natural language description of the initial scene.
The generated text should faithfully describe all the objects in the corresponding input frame.

\paragraph{Natural language explanations for sequences of pivotal events.}
Given a sequence of pivotal events for a simulation and the goal, generate a natural language description to explain the solution simulation.
The generated text should faithfully summarize the simulation by explaining the causal sequence of salient events in it. 

The goal of natural language generation for our task is to explain the pivotal physical events in the simulation so that an end user can solve the task more efficiently and reliably. Hence, we experimented with treating the physical event description generation as (1) \textbf{Table-to-Text Generation} and as (2) \textbf{Language Modeling}. The salient event detection component of our system serves as the {\em content selection} component of the natural language generation pipeline.
We describe the two approaches in the following sections.

\subsection{Table-to-Text Generation}
For the initial state description, the input is the structured table representation of the initial state, and the model generates a textual description conditioned on the input table.
Similarly, for the salient events explanation, the model produces the description given the structured table representation of all the salient events as the input.
Effective table-to-text generation can be leveraged to teach AI agents to solve tasks in natural language and output explanation for the steps in the task solution.

For both generation tasks, we use the model from \citet{puduppully2019entity} which is a neural model for table-to-text generation by explicitly modeling entities.\footnote{We also tried to use \citet{puduppully2019data}, but it requires a domain-specific relation extraction model to generate a specialized input, so we could not use it.}
Since our desired generations are ``entity coherent", in that their coherence depends on the introduction and discussion of entities in discourse \citep{karamanis2004evaluating}, the entity-based table-to-text generation model is a proper method for our task.
Unlike previous neural models treating entities as ordinary tokens, following \citet{puduppully2019entity}, we explicitly create entity representations for our objects in the physical environment and update their representation as the text is generated. 

The model input is a list of table records as $\{r_{j,l}\}_{l=1, j=1,\dots,|r|}^{L}$ where $|r|$ is the number of records for this example, and $L$ is the number of features for each record.
For example, $r_{j,1}$ are values and $r_{j,2}$ are entities.
The output $y$ is description with words $y=[y_1,\dots,y_{|y|}]$ where $|y|$ is the length of the description.

\paragraph{Encoder.}
We first create embeddings $\mathbf{r}_{j,l}$ of the features $r_{j,l}$, and then use a feed-forward layer to obtain the record embeddings $\mathbf{r}_{j}$.
\begin{equation*}
    \mathbf{r}_{j} = \text{ReLU}(\mathbf{W}_r[\mathbf{r}_{j,1},\dots,\mathbf{r}_{j,L}] + \mathbf{b}_r),
\end{equation*}
where $\mathbf{W}_r$ and $\mathbf{b}_r$ are model parameters.
From the record embeddings, we then use two methods to create the encoder outputs $\{\mathbf{e}_{j}\}_{j=1}^{|r|}$:
\begin{itemize}
    \item \textbf{AVG}. We use $\mathbf{e}_{j} = \mathbf{r}_{j}$, and the first hidden state of the decoder is the average of the record representations: $\text{avg}(\{\mathbf{e}_{j}\}_{j=1}^{|r|})$.
    \item \textbf{BiLSTM}. To account for the chronological order in the physical simulation, we use a BiLSTM over $[\mathbf{r}_{1},\dots,\mathbf{r}_{|r|}]$, whose hidden states are extracted as $\{\mathbf{e}_{j}\}_{j=1}^{|r|}$. 
The first hidden state of the decoder is initialized with the concatenation of the final step hidden states of the BiLSTM.
\end{itemize}

\paragraph{Entity memory.}
For each unique entity $k$ (i.e., one of $r_{j,2}$ values), we compute $\mathbf{x}_{k}$ as the average embeddings of all records which satisfy $r_{j,2}=k$.
During each decoding step $t$, we maintain an entity memory representation $\mathbf{u}_{t,k}$, and initialize it at $t=-1$ as:
\begin{equation*}
    \mathbf{u}_{t=-1,k} = \mathbf{W}_{i}\mathbf{x}_{k},
\end{equation*}
where $\mathbf{W}_{i}$ is a model parameter.

Denote the hidden state of the decoder at $t$ as $\mathbf{d}_{t}$.
We update the entity representation $\mathbf{u}_{k,t}$ at each $t$ with a gating mechanism as follows:
\begin{equation*}
\begin{split}
\boldsymbol{\gamma}_{t} & = \sigma(\mathbf{W}_{d}\mathbf{d}_{t}+\mathbf{b}_{d}), \\
    \boldsymbol{\delta}_{t,k} & = \boldsymbol{\gamma}_{t} \odot \sigma(\mathbf{W}_{e}\mathbf{d}_{t}+\mathbf{b}_{e} + \mathbf{W}_{f}\mathbf{u}_{t-1,k}+\mathbf{b}_{f}), \\
    \mathbf{\tilde{u}}_{t,k} & = \mathbf{W}_{g}\mathbf{d}_{t}, \\
    \mathbf{u}_{t,k} & = (1-\boldsymbol{\delta}_{t,k}) \odot \mathbf{u}_{t-1,k}  + \boldsymbol{\delta}_{t,k} \odot \mathbf{\tilde{u}}_{t,k},
\end{split}
\end{equation*}
where $\mathbf{W}_{d,e,f,g}$ and $\mathbf{b}_{d,e,f}$ are model parameters, and $\odot$ is element-wise product.
$\boldsymbol{\gamma}_{t}$ indicates if there should be an update at $t$, and $\boldsymbol{\delta}_{t,k}$ controls the update by interpolating between the previous $\mathbf{u}_{t-1,k}$ and candidate entity memory $\mathbf{\tilde{u}}_{t,k}$.

\paragraph{Hierarchical attention.}
We then use a hierarchical attention mechanism such that the decoder can first focus on entities and then the records for these entities.
We can rearrange the encoder output $\mathbf{e}_j$ in two-dimensional $\mathbf{g}_{k,z}$, where $k$ is index for entities and $z$ is the index for records of corresponding entities.
For each entity, we can compute the attention over its records along $z$, and compute the entity context vector $\mathbf{s}_{t,k}$:
\begin{equation*}
\begin{split}
  \alpha_{t,k,z}   & \propto \exp(\mathbf{d}_{t}^{\intercal}\mathbf{W}_{a}\mathbf{g}_{k,z}), \quad \sum_{z}\alpha_{t,k,z}=1, \\
  \mathbf{s}_{t,k} & = \sum_{z}\alpha_{t,k,z}\mathbf{g}_{k,z}.
\end{split}
\end{equation*}

Then we compute the higher level attention over entities along $k$, and compute the encoder context vector $\mathbf{q}_{t}$:
\begin{equation*}
\begin{split}
  \phi_{t,k}   & \propto \exp(\mathbf{d}_{t}^{\intercal}\mathbf{W}_{h}\mathbf{u}_{t,k}), \quad \sum_{k}\phi_{t,k}=1, \\
  \mathbf{q}_{t} & = \sum_{k}\phi_{t,k}\mathbf{s}_{t,k}.
\end{split}
\end{equation*}

\paragraph{Decoder.}
The encoder context vector $\mathbf{q}_{t}$ is then used in the decoder to compute a probability for each output token $y_t$:
\begin{equation*}
\begin{split}
  \mathbf{d}_{t}^{\text{att}} & = \tanh(\mathbf{W}_{c}[\mathbf{d}_{t};\mathbf{q}_{t}]), \\
  p_{\text{gen}}(y_t|y_{<t},r) & = \softmax_{y_t}(\mathbf{W}_{y}\mathbf{d}_{t}^{\text{att}} + \mathbf{b}_{y}).
\end{split}
\end{equation*}

In both generation tasks, we fine-tune the entity model provided by \citet{puduppully2019entity} for $125$ epochs. We use the same training hyperparameters and select the best model using token-match accuracy following \citet{puduppully2019entity}. 

\subsection{Language Modeling} 
We fine-tune a language model (LM) to generate descriptions of the initial state and explanations for sequences of pivotal physical events using the training split of our dataset. We use the pre-trained GPT-large \citep{radford2018improving} LM, which is a multi-layer transformer-based \citep{vaswani2017attention} model.

For the generation of initial state descriptions, the LM is fine-tuned conditioned on the objects (such as ball, jar, etc.) and their attributes (such as dynamic, static, color, size, etc.) extracted from the simulator described in Section~\ref{sec:representation} and the human written descriptions. So, the input context during training is defined as follows:
\begin{equation*}
C_{init} = o_1, o_2, \ldots, o_n, \textrm{``In the physical simulation ''}
\end{equation*}
where $o_1$, $o_2$, ..., $o_n$ is the list of extracted objects with their attributes, e.g., ``small red dynamic ball''. The model is trained to generate the initial scene description $s$ according to a conditional language modeling objective.  The objective is to maximize:
\begin{equation*}
\sum_i \log P(s_i | s_{i-k}, \ldots, s_{i-1}, C_{init}; \Theta),
\end{equation*}
where $k$ is the size of the context window (in our case $k$ is always greater than the length of $s$ so that the entire explanation is within the context). 
The conditional probability $P$ is modeled by a neural network with parameters $\Theta$ conditioned on $C_{init}$ and previous tokens.

For explanations of the salient physical events in the simulation, the LM is fine-tuned conditioned on the initial state descriptions and the human generated reasoning. So, the input context during training is defined as follows:
\begin{equation*}
C_{sim} = \textrm{``init\_scene. The red ball is placed and ''}
\end{equation*}
The model is trained to generate the physical reasoning $r$ by maximizing the following objective:
\[
\sum_i \log P(r_i | r_{i-k}, \ldots, r_{i-1}, C_{sim}; \Theta).
\]

We generate sequences of maximum length $40$, use a batch size of $12$, train for a maximum of $50$ epochs, selecting the best model based on validation BLEU and perplexity scores. The learning rate was set to $10^{-6}$, warmed up linearly with proportion $0.01$ and weight decay $0.01$. We experimented both with temperature $1.0$ and lower temperatures ($0.1$, $0.2$) to restrict generation to the physics domain and avoid diversity. For word sampling, we tried top $k$ as $3$ and $5$ as well as greedy ($k=1$). We found that the temperature of $0.1$ with $k=3$ worked best.

We note that it is not fair to compare the generated text by the table-to-text model and the LM because the input to the table-to-text model is structured with fine-grained details while the input to the LM is an unstructured prompt. A promising approach would be one that uses a table encoder with a pre-trained language model that is more robust and generalizable.

%% file: 04evaluation_metrics.tex
\section{Evaluation Metrics}
We evaluate our models using both automatic metrics and human evaluations.
\subsection{Automatic Metrics}
We use precision, recall, and F1 for the pivotal event classification task which can be formulated as a binary classification problem.

For the natural language description of initial frames and solution simulations, we use automatic metrics including BLEU-1, BLEU-2, ROUGE\_L, and METEOR using the implementation from \citet{sharma2017nlgeval}. %\footnote{\url{https://github.com/Maluuba/nlg-eval}}

\subsection{Human Evaluations}
The automated metrics for generation evaluation are very crude and do not measure the correctness and coverage of actual physical concepts or even the natural ordering in which physical events occur in a given simulation. For example, an object first falls and then it hits the ground or an object first falls on some other object which then causes the second object to be set in motion. So, we deployed human evaluations to measure the quality of the physical concepts captured by our language generation models in terms of \emph{validity} and \emph{coverage}.

To measure the \emph{validity} of initial scene descriptions, we showed humans the generated description for a task, the initial frames from that task, and three random distractor initial scenes from other tasks which may or may not be from the same template.
Then, we asked them to select the frame that belongs to the task being described. 
% Note that tasks within a template have the exact same objects and very similar sequences of events, the only difference being the initial position of the objects in the simulation. 
% It suffices to say that the generated text has to be very detailed and descriptive, e.g., ``red ball is $3/4$ of the way to the right boundary'', to be correctly matched to its simulation frame. 
This evaluates how faithful and accurate the generated description is to the input initial state. If the generated text does not include a detailed description of the objects, their attributes, and their positions, it would be difficult for humans to map them to the correct initial scene.

For evaluating the \emph{validity} of pivotal events descriptions, we showed humans the generated text for a task, the initial state of that task, and three distractor initial states generated from the same task but with positions of the red ball that do not solve the task. 
Then, we asked them to select the correct initial state with the red ball that would eventually reach the task goal.
A good simulation description should give higher accuracy for humans to choose the correct solution.
Note that we also evaluated the human generated initial state description and pivotal events description by asking annotators to match the human natural language descriptions that we collected and found the average accuracy to only be $70.2\%$ for the initial scene description and $44.7\%$ for the pivotal events description (Table~\ref{tab:human_eval_validity}). This is because of reporting bias, i.e., humans rarely state events that are obvious \citep{forbes2017verb}. For example, a falling ball would bounce multiple times or an object pushed off an elevated bar by another object would have a projectile motion. Lack of such fine-grained explanations is what makes the human evaluation of human generated descriptions especially for the sequence of pivotal events have poor accuracy.

The PHYRE tasks incorporate three physical concepts in every simulation --- gravity, collision, friction. So, to measure \emph{coverage}, we show humans just the natural language description of the simulation and ask them to select words that would imply any of the three concepts. For example, ``rolling" or ``slipping" would imply friction, ``falling" would imply gravity, ``hit" would imply collision, etc. We note that many physical concepts are very abstract and even difficult to be noticed visually, let alone describe in natural language. For example, moving objects slow down due to friction, but this physical concept is so innate that humans would not generally use words that imply friction to describe what they see. This metric gives us an overview of what degree of coverage the text generation models have for each of the three physical concepts.

For all our human evaluations we used MTurk and collected $3$ annotations per instance and report the majority. We paid Turkers $50$ cents per instance for the validity evaluation and $50$ cents per instance for the coverage evaluation.

\begin{table}[t!]
\centering
\resizebox{\columnwidth}{!}{%
\begin{tabular}{lccc}
\toprule
                    & Precision & Recall & F1 \\ 
\midrule
Positive            & 0.01      & 0.11   & 0.02 \\
Decision Tree       & 0.87      & 0.86   & 0.87 \\
MLP                 & 0.90      & 0.91   & 0.90 \\
\bottomrule
\end{tabular}
}
\caption{Results on pivotal events classification.}
\label{tab:results_salient}
\end{table}

\begin{table*}[t!]
\centering
\resizebox{\textwidth}{!}{%
\begin{tabular}{lcccc|cccc}
        \toprule
        & \multicolumn{4}{c}{Initial state description} & \multicolumn{4}{c}{Pivotal events description} \\ \hline
         & BLEU-1 & BLEU-2 & ROUGE\_L & METEOR & BLEU-1 & BLEU-2 & ROUGE\_L & METEOR \\ \midrule
GPT \citep{radford2018improving}       
& 15.37 & 2.25 & 20.08 & 9.93 & 24.32 & 3.89 & 26.82 & 12.14 \\
AVG \citep{puduppully2019entity}       
& 15.37 & 11.38 & 22.53 & 24.09 & 20.53 & 15.89 & 29.11 & 27.38 \\ 
BiLSTM \citep{puduppully2019entity}     
& 14.74 & 10.59 & 21.35 & 23.00 & 20.36 & 15.48 & 27.93 & 26.91  \\
\bottomrule
\end{tabular}
}
\caption{Automatic evaluation of initial state and pivotal events descriptions on the test set.}
\label{tab:automatic_eval_initial_sim}
\end{table*}

%% file: 05results_discussion.tex
\section{Experimental Results and Discussion}
Table \ref{tab:results_salient} summarizes the performance of the pivotal events classifiers.
The decision tree and MLP classifiers get very high performance with $0.87$ and $0.9$ F1 scores respectively. The baseline classifies every event as pivotal and thus performs very poorly.

From the decision tree, we extract feature importance values for each of the $13$ input variables described in Section~\ref{sec:methods}. The most important variable is the time step of the collision, with a weight of $0.178$. The most important features for classification were an object's collision position, its velocity, and then its angle of rotation. Given such strong results for identifying pivotal events, we were able to predict the salient events of previously unseen simulations and that helped in the next step of generation descriptions of salient events. 

\begin{table}[t!]
\centering
\resizebox{\columnwidth}{!}{%
\begin{tabular}{lcc}
\toprule
                                                        & Initial state & Pivotal events \\ \midrule
Random classifier                                       &   25.0        &   25.0     \\ \midrule
GPT \citep{radford2018improving}                        &   23.8        &   26.8     \\   
AVG \citep{puduppully2019entity}                        &   50.8        &   36.6     \\
BiLSTM \citep{puduppully2019entity}                     &   58.1        &   40.2     \\ \midrule
Human annotation                                        &   70.2        &   44.7     \\
\bottomrule
\end{tabular}
}
\caption{Human evaluation for \emph{validity} accuracy of initial state and simulation descriptions on test set.}
\label{tab:human_eval_validity}
\end{table}

\begin{table}[t!]
\centering
\scriptsize
\begin{tabular}{lccc}
\toprule
                                                        & Gravity & Friction & Collision  \\ \midrule
GPT \citep{radford2018improving}                        &  89.3   & 2.0      & 16.0       \\
AVG \citep{puduppully2019entity}                        &  100.4  & 61.6     & 71.8       \\
BiLSTM \citep{puduppully2019entity}                     &  99.2   & 70.7     & 71.1       \\
\midrule
Human annotation                                        &  96.7   & 43.0     & 57.0       \\
\bottomrule
\end{tabular}%
\caption{Human evaluation for \emph{coverage} accuracy of physical concepts in simulation descriptions on test set.}
\label{tab:human_eval_coverage}
\end{table}

\begin{table}[t!]
\centering
\scriptsize
\begin{tabularx}{\columnwidth}{l>{\raggedright}X}
\toprule
Input records & ... green\textbar green\_circle\_0\textbar OBJ\_COLOR\textbar \\
INITIAL\_STATE circle\textbar green\_circle\_0\textbar OBJ\_TYPE\textbar \\
INITIAL\_STATE dynamic\textbar green\_circle\_0\textbar OBJ\_STATE\textbar \\
INITIAL\_STATE 76\textbar green\_circle\_0\textbar X\textbar \\
INITIAL\_STATE 162\textbar green\_circle\_0\textbar Y\textbar \\
INITIAL\_STATE...\tabularnewline
\midrule
Gold annotation & The red and green balls fall. The red ball lands on the ground and the green ball lands on the red ball and rolls to the right over the black vertical bar.\tabularnewline
\midrule
Generation (AVG) & The red ball lands in the cubby and the green ball lands on top and a little to the right, sending the green ball right. It rolls over the short black wall of the cage and onto the floor, where it keeps rolling right towards the purple goal... \tabularnewline
Generation (BiLSTM) & The red ball falls and knocks the green ball off of its curved black platform and to the left. It rolls leftwards and continues falling until it lands on the purple floor...\tabularnewline
\bottomrule
\end{tabularx}
\caption{Example input records, gold annotation, and generated simulation description from the AVG and BiLSTM models, taken from example 00014:394. We show only a short segment of the actual input records.}
\label{tab:input_output} 
\end{table}

Table~\ref{tab:automatic_eval_initial_sim} shows the performance of the three text generation models using automatic metrics. The table-to-text models perform better than the language model on most of the metrics. The AVG model performs slightly better than the BiLSTM on both generation tasks. However, these metrics are a very crude measure of physical reasoning performance and are not intuitive. The human evaluations, on the other hand, are more informative and insightful.

\paragraph{Human evaluation -- validity.}
While the GPT model can achieve scores comparable to the data-to-text models using automated metrics, its performance using human evaluation is as good as chance, as shown in Table~\ref{tab:human_eval_validity}. We found that the GPT LM generation was very high-level and is not useful for humans to decide which tasks (among the correct and distractor choices) the generated solution explanation of the initial state and pivotal events match.
By contrast, AVG and BiLSTM have significantly higher accuracy, mainly because their output is more fine-grained and so gives a more thorough explanation of the solution.
Surprisingly, the human annotations of the descriptions that we collected as ground truth are not perfect either, indicating that humans tend to produce sentences that are not sufficiently discriminate and even sometimes skip obvious details such as whether the ball rolls to the left vs. right.

\paragraph{Human evaluation -- coverage.}
Table \ref{tab:human_eval_coverage} shows the results for coverage of physical concepts.
The outputs of the GPT model are repetitive and not grammatical, containing little explanation of physical concepts.
AVG and BiLSTM, on the other hand, can generate text that contains fine-grained descriptions of physical concepts even sometimes better than those generated by humans.
This is because humans don't describe everyday commonsense concepts using fine-grained language, while the AVG and BiLSTM models tend to generate long detailed descriptions containing various words for gravity (e.g., falls, drop, slope, land), friction (e.g., roll, slide, trap, travel, stuck, remain), and collision (e.g., hit, collide, impact, land, pin, bounce).

\paragraph{Qualitative analysis.}
An example of the model inputs and outputs is in Table \ref{tab:input_output} and taken from simulation id 00014:394. Here we make two observations. 
First, the generated descriptions are not as succinct as the gold annotations, because our model is obtained from fine-tuning an entity-based model pre-trained on generating long Rotowire game summaries~\citep{wiseman-etal-2017-challenges}. 
Second, the output generated by the BiLSTM model predicts the incorrect direction of motion for the green ball, an error that is occasionally seen across generation descriptions of both models. This indicates that a table-to-text paradigm for generating such solution explanations is not adequate for learning the direction of motion for the physical reasoning required for these explanations.

%% file: 06related_work.tex
\section{Related Work}
\paragraph{Qualitative physics and Visual reasoning.}
Qualitative physics aims to represent and reason about the behavior of physical systems \citep{forbus1988qualitative}.
\citet{mccloskey1983naive,mccloskey1983intuitive} suggests that people use simplified intuitive theories to understand the physical world in day-to-day life.
Earlier works explored using probabilistic simulations to train physical inference through physical simulations \citep{battaglia2013simulation,zhang2016comparative}.
Recent papers use neural networks over visual inputs to predict future pixels \citep{finn2016unsupervised,lerer2016learning,mirza2016generalizable,du2019task} or make qualitative predictions \citep{groth2018shapestacks,li2016fall,li2017visual,janner2018reasoning,wu2015galileo,Mao2019NeuroSymbolic}.
Furthermore, several frameworks and benchmarks have been introduced to test visual reasoning such as PHYRE \citep{bakhtin2019phyre}, Mujoco \citep{todorov2012mujoco}, and Intphys \citep{riochet2018intphys}, some of which are combined with natural language for question answering such as NLVR \citep{suhr2017corpus}, CLEVR \citep{johnson2017clevr}, and VQA \citep{antol2015vqa}.
In a parallel work, \citet{yi2020clevrer} introduced the CLEVRER dataset for reasoning about collision events from videos with different types of questions.
In contrast, we develop the ability to reason and explain the behavior of dynamic physical systems by generating natural language.

\paragraph{Natural language explanations and Commonsense reasoning.}
Several recent papers propose to use natural language for explanation and commonsense reasoning \citep{lei2016rationalizing,camburu2018snli,forbes2017verb,chai2018language,forbes2019neural,rajani2019explain,deyoung2019eraser}.
\citet{lei2016rationalizing}, for example, generate textual rationales for sentiment analysis by highlighting phrases in the input.
\citet{forbes2017verb} learn the physical knowledge of actions and objects from natural language.
\citet{camburu2018snli} propose e-SNLI by generating explanations for the natural language inference problem at a cost of performance.
\citet{rajani2019explain} propose to use LMs to generate explanations that can be used during training and inference in a classifier and significantly improve CommonsenseQA performance. 
\citet{bisk2019piqa} propose to use a question answering task to test the model's physical commonsense and reasoning ability. 
In contrast to the previous work, we focus on identifying pivotal physical events and then generating natural language explanations for them. We find that this two-step approach works more effectively.

\paragraph{Table-to-text generation.}
Table-to-text generation aims to produce natural language output from structured input.
Applications include generating sports commentaries from game records \citep{tanaka1998reactive,chen2008learning,taniguchi2019generating}, weather forecasts \citep{liang2009learning,konstas2012unsupervised,mei2015talk}, biographical texts from Wikipedia infoboxes \citep{lebret2016neural,sha2018order,liu2018table,perez2018bootstrapping}, descriptions of knowledge bases \citep{o2000optimising,trisedya-etal-2018-gtr,zhu2019triple,yu2019cosql} and source code \citep{iyer2016summarizing}, and dialog response generation from slot-value pairs \citep{wen-etal-2015-semantically}.

Recently, neural encoder-decoder models \citep{sutskever2014sequence,cho2014learning} based on attention \citep{bahdanau2014neural,luong2015effective} and copy mechanisms \citep{gu2016incorporating,gulcehre2016pointing} have shown promising results on table-to-text tasks \citep{wiseman-etal-2017-challenges,gehrmann-etal-2018-end,puduppully2019data,puduppully2019entity,iso-etal-2019-learning,castro-ferreira-etal-2019-neural,zhao2020bridging,chen2020logical}.
While traditional methods use different modules for each generation stage in a pipeline \citep{reiter2000building}, neural table-to-text models are trained on large-scale datasets, relying on representation learning for generating coherent and grammatical texts.
\citet{puduppully2019data} propose a neural network approach that first selects data records to be mentioned and then generates a summary from the selected data, in an end-to-end fashion.
\citet{chen2020few} use pre-trained language models to generate descriptions for tabular data in a few shot setting.

%% file: 07conclusion.tex
\section{Conclusions and Future Directions}
ESPRIT uses a two-step approach for qualitative physical reasoning.
To train models that can describe physical tasks, we collected open-ended natural language text descriptions of initial states and pivotal physical events in a 2D simulation from human annotators.
We then trained a model to identify these pivotal events and then fine-tuned on pre-trained table-to-text generation and language models without using the image representations of the actual simulation frames. 

Our results indicate that table-to-text models perform better than language models on generating valid explanations of physical events but there is a lot more room for improvement compared to human annotations. 
We hope that the dataset we collected will facilitate research in using natural language for physical reasoning.

Reinforcement Learning (RL) agents may be able to solve physical tasks much more efficiently by leveraging natural language reasoning as opposed to model-free approaches that are often highly sample-inefficient.
An RL agent that leverages natural language descriptions of physical events to reason about the solution for a given goal (similar to \citet{zhong2019rtfm}) or for reward shaping (similar to \citet{ijcai2019-331}) could be a compelling line of future research.

More importantly, having a model that can meaningfully reason about commonsense qualitative physics could be interpretable and more robust, as they might focus on the parts of physical dynamics that are relevant for generalization to new scenarios. 
Such systems are widely applicable to self-driving cars or tasks that involve human-AI interactions, such as robots performing everyday human tasks like making coffee or even collaboratively helping with rescue operations.

%% file: 08acknowledgment.tex
\section*{Acknowledgments}
We would like to thank Abhinand Sivaprasad for his helpful discussions and annotations.
We also thank the anonymous reviewers for their feedback.

%% file: acl2020.bbl
\begin{thebibliography}{68}
\expandafter\ifx\csname natexlab\endcsname\relax\def\natexlab#1{#1}\fi

\bibitem[{Antol et~al.(2015)Antol, Agrawal, Lu, Mitchell, Batra,
  Lawrence~Zitnick, and Parikh}]{antol2015vqa}
Stanislaw Antol, Aishwarya Agrawal, Jiasen Lu, Margaret Mitchell, Dhruv Batra,
  C~Lawrence~Zitnick, and Devi Parikh. 2015.
\newblock {VQA}: Visual question answering.
\newblock In \emph{ICCV}.

\bibitem[{Bahdanau et~al.(2015)Bahdanau, Cho, and Bengio}]{bahdanau2014neural}
Dzmitry Bahdanau, Kyunghyun Cho, and Yoshua Bengio. 2015.
\newblock Neural machine translation by jointly learning to align and
  translate.
\newblock In \emph{ICLR}.

\bibitem[{Bakhtin et~al.(2019)Bakhtin, van~der Maaten, Johnson, Gustafson, and
  Girshick}]{bakhtin2019phyre}
Anton Bakhtin, Laurens van~der Maaten, Justin Johnson, Laura Gustafson, and
  Ross Girshick. 2019.
\newblock {PHYRE}: A new benchmark for physical reasoning.
\newblock In \emph{NeurIPS}.

\bibitem[{Battaglia et~al.(2013)Battaglia, Hamrick, and
  Tenenbaum}]{battaglia2013simulation}
Peter~W Battaglia, Jessica~B Hamrick, and Joshua~B Tenenbaum. 2013.
\newblock Simulation as an engine of physical scene understanding.
\newblock \emph{Proceedings of the National Academy of Sciences},
  110(45):18327--18332.

\bibitem[{Bisk et~al.(2020)Bisk, Zellers, Bras, Gao, and Choi}]{bisk2019piqa}
Yonatan Bisk, Rowan Zellers, Ronan~Le Bras, Jianfeng Gao, and Yejin Choi. 2020.
\newblock {PIQA}: Reasoning about physical commonsense in natural language.
\newblock In \emph{AAAI}.

\bibitem[{Camburu et~al.(2018)Camburu, Rockt{\"a}schel, Lukasiewicz, and
  Blunsom}]{camburu2018snli}
Oana-Maria Camburu, Tim Rockt{\"a}schel, Thomas Lukasiewicz, and Phil Blunsom.
  2018.
\newblock e-{SNLI}: natural language inference with natural language
  explanations.
\newblock In \emph{NeurIPS}.

\bibitem[{Castro~Ferreira et~al.(2019)Castro~Ferreira, van~der Lee, van
  Miltenburg, and Krahmer}]{castro-ferreira-etal-2019-neural}
Thiago Castro~Ferreira, Chris van~der Lee, Emiel van Miltenburg, and Emiel
  Krahmer. 2019.
\newblock Neural data-to-text generation: A comparison between pipeline and
  end-to-end architectures.
\newblock In \emph{{EMNLP-IJCNLP}}.

\bibitem[{Chai et~al.(2018)Chai, Gao, She, Yang, Saba-Sadiya, and
  Xu}]{chai2018language}
Joyce~Y Chai, Qiaozi Gao, Lanbo She, Shaohua Yang, Sari Saba-Sadiya, and
  Guangyue Xu. 2018.
\newblock Language to action: Towards interactive task learning with physical
  agents.
\newblock In \emph{AAMAS}.

\bibitem[{Chen and Mooney(2008)}]{chen2008learning}
David~L Chen and Raymond~J Mooney. 2008.
\newblock Learning to sportscast: a test of grounded language acquisition.
\newblock In \emph{ICML}.

\bibitem[{Chen et~al.(2020{\natexlab{a}})Chen, Chen, Su, Chen, and
  Wang}]{chen2020logical}
Wenhu Chen, Jianshu Chen, Yu~Su, Zhiyu Chen, and William~Yang Wang.
  2020{\natexlab{a}}.
\newblock Logical natural language generation from open-domain tables.
\newblock In \emph{ACL}.

\bibitem[{Chen et~al.(2020{\natexlab{b}})Chen, Eavani, Chen, Liu, and
  Wang}]{chen2020few}
Zhiyu Chen, Harini Eavani, Wenhu Chen, Yinyin Liu, and William~Yang Wang.
  2020{\natexlab{b}}.
\newblock Few-shot {NLG} with pre-trained language model.
\newblock In \emph{ACL}.

\bibitem[{Cho et~al.(2014)Cho, van Merri{\"e}nboer, Gulcehre, Bahdanau,
  Bougares, Schwenk, and Bengio}]{cho2014learning}
Kyunghyun Cho, Bart van Merri{\"e}nboer, Caglar Gulcehre, Dzmitry Bahdanau,
  Fethi Bougares, Holger Schwenk, and Yoshua Bengio. 2014.
\newblock Learning phrase representations using {RNN} encoder{--}decoder for
  statistical machine translation.
\newblock In \emph{{EMNLP}}.

\bibitem[{DeYoung et~al.(2020)DeYoung, Jain, Rajani, Lehman, Xiong, Socher, and
  Wallace}]{deyoung2019eraser}
Jay DeYoung, Sarthak Jain, Nazneen~Fatema Rajani, Eric Lehman, Caiming Xiong,
  Richard Socher, and Byron~C Wallace. 2020.
\newblock {ERASER}: A benchmark to evaluate rationalized nlp models.
\newblock In \emph{ACL}.

\bibitem[{Du and Narasimhan(2019)}]{du2019task}
Yilun Du and Karthik Narasimhan. 2019.
\newblock Task-agnostic dynamics priors for deep reinforcement learning.
\newblock In \emph{ICML}.

\bibitem[{Finn et~al.(2016)Finn, Goodfellow, and Levine}]{finn2016unsupervised}
Chelsea Finn, Ian Goodfellow, and Sergey Levine. 2016.
\newblock Unsupervised learning for physical interaction through video
  prediction.
\newblock In \emph{NeurIPS}.

\bibitem[{Forbes and Choi(2017)}]{forbes2017verb}
Maxwell Forbes and Yejin Choi. 2017.
\newblock Verb physics: Relative physical knowledge of actions and objects.
\newblock In \emph{ACL}.

\bibitem[{Forbes et~al.(2019)Forbes, Holtzman, and Choi}]{forbes2019neural}
Maxwell Forbes, Ari Holtzman, and Yejin Choi. 2019.
\newblock Do neural language representations learn physical commonsense?
\newblock In \emph{{CogSci}}.

\bibitem[{Forbus(1988)}]{forbus1988qualitative}
Kenneth~D Forbus. 1988.
\newblock Qualitative physics: Past, present, and future.
\newblock In \emph{Exploring artificial intelligence}, pages 239--296.
  Elsevier.

\bibitem[{Gehrmann et~al.(2018)Gehrmann, Dai, Elder, and
  Rush}]{gehrmann-etal-2018-end}
Sebastian Gehrmann, Falcon Dai, Henry Elder, and Alexander Rush. 2018.
\newblock End-to-end content and plan selection for data-to-text generation.
\newblock In \emph{INLG}.

\bibitem[{Goyal et~al.(2019)Goyal, Niekum, and Mooney}]{ijcai2019-331}
Prasoon Goyal, Scott Niekum, and Raymond~J. Mooney. 2019.
\newblock Using natural language for reward shaping in reinforcement learning.
\newblock In \emph{IJCAI}.

\bibitem[{Groth et~al.(2018)Groth, Fuchs, Posner, and
  Vedaldi}]{groth2018shapestacks}
Oliver Groth, Fabian~B Fuchs, Ingmar Posner, and Andrea Vedaldi. 2018.
\newblock Shape{S}tacks: Learning vision-based physical intuition for
  generalised object stacking.
\newblock In \emph{ECCV}.

\bibitem[{Gu et~al.(2016)Gu, Lu, Li, and Li}]{gu2016incorporating}
Jiatao Gu, Zhengdong Lu, Hang Li, and Victor~O.K. Li. 2016.
\newblock Incorporating copying mechanism in sequence-to-sequence learning.
\newblock In \emph{ACL}.

\bibitem[{Gulcehre et~al.(2016)Gulcehre, Ahn, Nallapati, Zhou, and
  Bengio}]{gulcehre2016pointing}
Caglar Gulcehre, Sungjin Ahn, Ramesh Nallapati, Bowen Zhou, and Yoshua Bengio.
  2016.
\newblock Pointing the unknown words.
\newblock In \emph{ACL}.

\bibitem[{Iso et~al.(2019)Iso, Uehara, Ishigaki, Noji, Aramaki, Kobayashi,
  Miyao, Okazaki, and Takamura}]{iso-etal-2019-learning}
Hayate Iso, Yui Uehara, Tatsuya Ishigaki, Hiroshi Noji, Eiji Aramaki, Ichiro
  Kobayashi, Yusuke Miyao, Naoaki Okazaki, and Hiroya Takamura. 2019.
\newblock Learning to select, track, and generate for data-to-text.
\newblock In \emph{ACL}.

\bibitem[{Iyer et~al.(2016)Iyer, Konstas, Cheung, and
  Zettlemoyer}]{iyer2016summarizing}
Srinivasan Iyer, Ioannis Konstas, Alvin Cheung, and Luke Zettlemoyer. 2016.
\newblock Summarizing source code using a neural attention model.
\newblock In \emph{ACL}.

\bibitem[{Janner et~al.(2019)Janner, Levine, Freeman, Tenenbaum, Finn, and
  Wu}]{janner2018reasoning}
Michael Janner, Sergey Levine, William~T Freeman, Joshua~B Tenenbaum, Chelsea
  Finn, and Jiajun Wu. 2019.
\newblock Reasoning about physical interactions with object-oriented prediction
  and planning.
\newblock In \emph{ICLR}.

\bibitem[{Johnson et~al.(2017)Johnson, Hariharan, van~der Maaten, Fei-Fei,
  Lawrence~Zitnick, and Girshick}]{johnson2017clevr}
Justin Johnson, Bharath Hariharan, Laurens van~der Maaten, Li~Fei-Fei,
  C~Lawrence~Zitnick, and Ross Girshick. 2017.
\newblock {CLEVR}: A diagnostic dataset for compositional language and
  elementary visual reasoning.
\newblock In \emph{CVPR}.

\bibitem[{Karamanis et~al.(2004)Karamanis, Poesio, Mellish, and
  Oberlander}]{karamanis2004evaluating}
Nikiforos Karamanis, Massimo Poesio, Chris Mellish, and Jon Oberlander. 2004.
\newblock Evaluating centering-based metrics of coherence.
\newblock In \emph{ACL}.

\bibitem[{Konstas and Lapata(2012)}]{konstas2012unsupervised}
Ioannis Konstas and Mirella Lapata. 2012.
\newblock Unsupervised concept-to-text generation with hypergraphs.
\newblock In \emph{NAACL}.

\bibitem[{Lebret et~al.(2016)Lebret, Grangier, and Auli}]{lebret2016neural}
R{\'e}mi Lebret, David Grangier, and Michael Auli. 2016.
\newblock Neural text generation from structured data with application to the
  biography domain.
\newblock In \emph{EMNLP}.

\bibitem[{Lei et~al.(2016)Lei, Barzilay, and Jaakkola}]{lei2016rationalizing}
Tao Lei, Regina Barzilay, and Tommi Jaakkola. 2016.
\newblock Rationalizing neural predictions.
\newblock In \emph{EMNLP}.

\bibitem[{Lerer et~al.(2016)Lerer, Gross, and Fergus}]{lerer2016learning}
Adam Lerer, Sam Gross, and Rob Fergus. 2016.
\newblock Learning physical intuition of block towers by example.
\newblock In \emph{ICML}.

\bibitem[{Li et~al.(2016)Li, Azimi, Leonardis, and Fritz}]{li2016fall}
Wenbin Li, Seyedmajid Azimi, Ale{\v{s}} Leonardis, and Mario Fritz. 2016.
\newblock To fall or not to fall: A visual approach to physical stability
  prediction.
\newblock \emph{arXiv preprint arXiv:1604.00066}.

\bibitem[{Li et~al.(2017)Li, Leonardis, and Fritz}]{li2017visual}
Wenbin Li, Ales Leonardis, and Mario Fritz. 2017.
\newblock Visual stability prediction and its application to manipulation.
\newblock In \emph{2017 AAAI Spring Symposium Series}.

\bibitem[{Liang et~al.(2009)Liang, Jordan, and Klein}]{liang2009learning}
Percy Liang, Michael~I Jordan, and Dan Klein. 2009.
\newblock Learning semantic correspondences with less supervision.
\newblock In \emph{ACL-IJCNLP}.

\bibitem[{Liu et~al.(2018)Liu, Wang, Sha, Chang, and Sui}]{liu2018table}
Tianyu Liu, Kexiang Wang, Lei Sha, Baobao Chang, and Zhifang Sui. 2018.
\newblock Table-to-text generation by structure-aware seq2seq learning.
\newblock In \emph{AAAI}.

\bibitem[{Luong et~al.(2015)Luong, Pham, and Manning}]{luong2015effective}
Thang Luong, Hieu Pham, and Christopher~D. Manning. 2015.
\newblock Effective approaches to attention-based neural machine translation.
\newblock In \emph{EMNLP}.

\bibitem[{Mao et~al.(2019)Mao, Gan, Kohli, Tenenbaum, and
  Wu}]{Mao2019NeuroSymbolic}
Jiayuan Mao, Chuang Gan, Pushmeet Kohli, Joshua~B Tenenbaum, and Jiajun Wu.
  2019.
\newblock The neuro-symbolic concept learner: Interpreting scenes, words, and
  sentences from natural supervision.
\newblock In \emph{ICLR}.

\bibitem[{McCloskey and Kohl(1983)}]{mccloskey1983naive}
Michael McCloskey and Deborah Kohl. 1983.
\newblock Naive physics: The curvilinear impetus principle and its role in
  interactions with moving objects.
\newblock \emph{Journal of Experimental Psychology: Learning, Memory, and
  Cognition}, 9(1):146.

\bibitem[{McCloskey et~al.(1983)McCloskey, Washburn, and
  Felch}]{mccloskey1983intuitive}
Michael McCloskey, Allyson Washburn, and Linda Felch. 1983.
\newblock Intuitive physics: the straight-down belief and its origin.
\newblock \emph{Journal of Experimental Psychology: Learning, Memory, and
  Cognition}, 9(4):636.

\bibitem[{Mei et~al.(2016)Mei, Bansal, and Walter}]{mei2015talk}
Hongyuan Mei, Mohit Bansal, and Matthew~R. Walter. 2016.
\newblock What to talk about and how? selective generation using {LSTM}s with
  coarse-to-fine alignment.
\newblock In \emph{NAACL}.

\bibitem[{Mirza et~al.(2016)Mirza, Courville, and
  Bengio}]{mirza2016generalizable}
Mehdi Mirza, Aaron Courville, and Yoshua Bengio. 2016.
\newblock Generalizable features from unsupervised learning.
\newblock \emph{arXiv preprint arXiv:1612.03809}.

\bibitem[{O’Donnell et~al.(2000)O’Donnell, Knott, Oberlander, and
  Mellish}]{o2000optimising}
Michael O’Donnell, Alistair Knott, Jon Oberlander, and Chris Mellish. 2000.
\newblock Optimising text quality in generation from relational databases.
\newblock In \emph{INLG}.

\bibitem[{Perez-Beltrachini and Lapata(2018)}]{perez2018bootstrapping}
Laura Perez-Beltrachini and Mirella Lapata. 2018.
\newblock Bootstrapping generators from noisy data.
\newblock In \emph{NAACL}.

\bibitem[{Puduppully et~al.(2019{\natexlab{a}})Puduppully, Dong, and
  Lapata}]{puduppully2019data}
Ratish Puduppully, Li~Dong, and Mirella Lapata. 2019{\natexlab{a}}.
\newblock Data-to-text generation with content selection and planning.
\newblock In \emph{AAAI}.

\bibitem[{Puduppully et~al.(2019{\natexlab{b}})Puduppully, Dong, and
  Lapata}]{puduppully2019entity}
Ratish Puduppully, Li~Dong, and Mirella Lapata. 2019{\natexlab{b}}.
\newblock Data-to-text generation with entity modeling.
\newblock In \emph{ACL}.

\bibitem[{Radford et~al.(2018)Radford, Narasimhan, Salimans, and
  Sutskever}]{radford2018improving}
Alec Radford, Karthik Narasimhan, Tim Salimans, and Ilya Sutskever. 2018.
\newblock Improving language understanding by generative pre-training.
\newblock Technical report, OpenAI.

\bibitem[{Rajani et~al.(2019)Rajani, McCann, Xiong, and
  Socher}]{rajani2019explain}
Nazneen~Fatema Rajani, Bryan McCann, Caiming Xiong, and Richard Socher. 2019.
\newblock Explain yourself! leveraging language models for commonsense
  reasoning.
\newblock In \emph{ACL}.

\bibitem[{Reiter and Dale(2000)}]{reiter2000building}
Ehud Reiter and Robert Dale. 2000.
\newblock \emph{Building natural language generation systems}.
\newblock Cambridge university press.

\bibitem[{Riochet et~al.(2018)Riochet, Castro, Bernard, Lerer, Fergus, Izard,
  and Dupoux}]{riochet2018intphys}
Ronan Riochet, Mario~Ynocente Castro, Mathieu Bernard, Adam Lerer, Rob Fergus,
  V{\'e}ronique Izard, and Emmanuel Dupoux. 2018.
\newblock Int{P}hys: A framework and benchmark for visual intuitive physics
  reasoning.
\newblock \emph{arXiv preprint arXiv:1803.07616}.

\bibitem[{Sha et~al.(2018)Sha, Mou, Liu, Poupart, Li, Chang, and
  Sui}]{sha2018order}
Lei Sha, Lili Mou, Tianyu Liu, Pascal Poupart, Sujian Li, Baobao Chang, and
  Zhifang Sui. 2018.
\newblock Order-planning neural text generation from structured data.
\newblock In \emph{AAAI}.

\bibitem[{Sharma et~al.(2017)Sharma, Asri, Schulz, and
  Zumer}]{sharma2017nlgeval}
Shikhar Sharma, Layla~El Asri, Hannes Schulz, and Jeremie Zumer. 2017.
\newblock Relevance of unsupervised metrics in task-oriented dialogue for
  evaluating natural language generation.
\newblock \emph{arXiv preprint arXiv:1706.09799}.

\bibitem[{Suhr et~al.(2017)Suhr, Lewis, Yeh, and Artzi}]{suhr2017corpus}
Alane Suhr, Mike Lewis, James Yeh, and Yoav Artzi. 2017.
\newblock A corpus of natural language for visual reasoning.
\newblock In \emph{ACL}.

\bibitem[{Sutskever et~al.(2014)Sutskever, Vinyals, and
  Le}]{sutskever2014sequence}
Ilya Sutskever, Oriol Vinyals, and Quoc~V Le. 2014.
\newblock Sequence to sequence learning with neural networks.
\newblock In \emph{NeurIPS}.

\bibitem[{Tanaka-Ishii et~al.(1998)Tanaka-Ishii, Hasida, and
  Noda}]{tanaka1998reactive}
Kumiko Tanaka-Ishii, K{\^o}iti Hasida, and Itsuki Noda. 1998.
\newblock Reactive content selection in the generation of real-time soccer
  commentary.
\newblock In \emph{COLING}.

\bibitem[{Taniguchi et~al.(2019)Taniguchi, Feng, Takamura, and
  Okumura}]{taniguchi2019generating}
Yasufumi Taniguchi, Yukun Feng, Hiroya Takamura, and Manabu Okumura. 2019.
\newblock Generating live soccer-match commentary from play data.
\newblock In \emph{AAAI}.

\bibitem[{Todorov et~al.(2012)Todorov, Erez, and Tassa}]{todorov2012mujoco}
Emanuel Todorov, Tom Erez, and Yuval Tassa. 2012.
\newblock Mu{J}o{C}o: A physics engine for model-based control.
\newblock In \emph{2012 IEEE/RSJ International Conference on Intelligent Robots
  and Systems}. IEEE.

\bibitem[{Trisedya et~al.(2018)Trisedya, Qi, Zhang, and
  Wang}]{trisedya-etal-2018-gtr}
Bayu~Distiawan Trisedya, Jianzhong Qi, Rui Zhang, and Wei Wang. 2018.
\newblock {GTR}-{LSTM}: A triple encoder for sentence generation from {RDF}
  data.
\newblock In \emph{ACL}.

\bibitem[{Vaswani et~al.(2017)Vaswani, Shazeer, Parmar, Uszkoreit, Jones,
  Gomez, Kaiser, and Polosukhin}]{vaswani2017attention}
Ashish Vaswani, Noam Shazeer, Niki Parmar, Jakob Uszkoreit, Llion Jones,
  Aidan~N Gomez, {\L}ukasz Kaiser, and Illia Polosukhin. 2017.
\newblock Attention is all you need.
\newblock In \emph{NeurIPS}.

\bibitem[{Wen et~al.(2015)Wen, Ga{\v{s}}i{\'c}, Mrk{\v{s}}i{\'c}, Su, Vandyke,
  and Young}]{wen-etal-2015-semantically}
Tsung-Hsien Wen, Milica Ga{\v{s}}i{\'c}, Nikola Mrk{\v{s}}i{\'c}, Pei-Hao Su,
  David Vandyke, and Steve Young. 2015.
\newblock Semantically conditioned {LSTM}-based natural language generation for
  spoken dialogue systems.
\newblock In \emph{EMNLP}.

\bibitem[{Wiseman et~al.(2017)Wiseman, Shieber, and
  Rush}]{wiseman-etal-2017-challenges}
Sam Wiseman, Stuart Shieber, and Alexander Rush. 2017.
\newblock Challenges in data-to-document generation.
\newblock In \emph{EMNLP}.

\bibitem[{Wu et~al.(2015)Wu, Yildirim, Lim, Freeman, and
  Tenenbaum}]{wu2015galileo}
Jiajun Wu, Ilker Yildirim, Joseph~J Lim, Bill Freeman, and Josh Tenenbaum.
  2015.
\newblock Galileo: Perceiving physical object properties by integrating a
  physics engine with deep learning.
\newblock In \emph{NeurIPS}.

\bibitem[{Yi et~al.(2020)Yi, Gan, Li, Kohli, Wu, Torralba, and
  Tenenbaum}]{yi2020clevrer}
Kexin Yi, Chuang Gan, Yunzhu Li, Pushmeet Kohli, Jiajun Wu, Antonio Torralba,
  and Joshua~B Tenenbaum. 2020.
\newblock {CLEVRER}: Collision events for video representation and reasoning.
\newblock In \emph{ICLR}.

\bibitem[{Yu et~al.(2019)Yu, Zhang, Er, Li, Xue, Pang, Lin, Tan, Shi, Li,
  Jiang, Yasunaga, Shim, Chen, Fabbri, Li, Chen, Zhang, Dixit, Zhang, Xiong,
  Socher, Lasecki, and Radev}]{yu2019cosql}
Tao Yu, Rui Zhang, Heyang Er, Suyi Li, Eric Xue, Bo~Pang, Xi~Victoria Lin,
  Yi~Chern Tan, Tianze Shi, Zihan Li, Youxuan Jiang, Michihiro Yasunaga,
  Sungrok Shim, Tao Chen, Alexander Fabbri, Zifan Li, Luyao Chen, Yuwen Zhang,
  Shreya Dixit, Vincent Zhang, Caiming Xiong, Richard Socher, Walter Lasecki,
  and Dragomir Radev. 2019.
\newblock {C}o{SQL}: A conversational text-to-{SQL} challenge towards
  cross-domain natural language interfaces to databases.
\newblock In \emph{{EMNLP-IJCNLP}}.

\bibitem[{Zhang et~al.(2016)Zhang, Wu, Zhang, Freeman, and
  Tenenbaum}]{zhang2016comparative}
Renqiao Zhang, Jiajun Wu, Chengkai Zhang, William~T Freeman, and Joshua~B
  Tenenbaum. 2016.
\newblock A comparative evaluation of approximate probabilistic simulation and
  deep neural networks as accounts of human physical scene understanding.
\newblock In \emph{CogSci}.

\bibitem[{Zhao et~al.(2020)Zhao, Walker, and Chaturvedi}]{zhao2020bridging}
Chao Zhao, Marilyn Walker, and Snigdha Chaturvedi. 2020.
\newblock Bridging the structural gap between encoding and decoding for
  data-to-text generation.
\newblock In \emph{ACL}.

\bibitem[{Zhong et~al.(2020)Zhong, Rockt{\"a}schel, and
  Grefenstette}]{zhong2019rtfm}
Victor Zhong, Tim Rockt{\"a}schel, and Edward Grefenstette. 2020.
\newblock {RTFM}: Generalising to new environment dynamics via reading.
\newblock In \emph{ICLR}.

\bibitem[{Zhu et~al.(2019)Zhu, Wan, Zhou, Chen, Qiu, Zhang, Jiang, and
  Yu}]{zhu2019triple}
Yaoming Zhu, Juncheng Wan, Zhiming Zhou, Liheng Chen, Lin Qiu, Weinan Zhang,
  Xin Jiang, and Yong Yu. 2019.
\newblock Triple-to-text: Converting rdf triples into high-quality natural
  languages via optimizing an inverse kl divergence.
\newblock In \emph{SIGIR}.

\end{thebibliography}
